\documentclass[11pt,a4paper]{article}
\usepackage{times,latexsym}
\usepackage{url}
\usepackage[T1]{fontenc}
\usepackage{graphicx}
\usepackage[inline]{enumitem}
\usepackage{makecell}

\usepackage[acceptedWithA]{tacl2021v1}

\usepackage{booktabs}
\usepackage{algorithm}
\usepackage{algorithmic}
\usepackage{amsmath, nccmath}
\usepackage{cleveref}
\usepackage[switch]{lineno}
\usepackage{scrextend}
\usepackage{xcolor}
\usepackage{color, colortbl}

\usepackage{xspace,mfirstuc,tabulary}

\newif\iftaclinstructions
\taclinstructionsfalse 
\iftaclinstructions
\renewcommand{\confidential}{}
\renewcommand{\anonsubtext}{(No author info supplied here, for consistency with
TACL-submission anonymization requirements)}
\newcommand{\instr}
\fi

\iftaclpubformat 

\else

\fi

\title{ARN: Analogical Reasoning on Narratives}

\author{
  Zhivar Sourati$^\diamond$
  \
  Filip Ilievski$^\dagger$
  \
  Pia Sommerauer$^*$
  \
  Yifan Jiang$^\diamond$
  \\
  \ \\
  $^\diamond$Information Sciences Institute, University of Southern California, Marina del Rey, CA, USA
  \\
  Department of Computer Science, University of Southern California, Los Angeles, CA, USA
  \ \\
  $^\dagger$Department of Computer Science, Vrije Universiteit Amsterdam, Netherlands
  \ \\
  $^*$Computational Linguistics \& Text Mining Lab, Vrije Universiteit Amsterdam, Netherlands
  \\
  \texttt{\{souratih,yifjia\}@isi.edu,\{f.ilievski,pia.sommerauer\}@vu.nl}
}

\date{}

\begin{document}
\maketitle

\begin{abstract}
As a core cognitive skill that enables the transferability of information across domains, analogical reasoning has been extensively studied for both humans and computational models. However, while cognitive theories of analogy often focus on narratives and study the distinction between surface, relational, and system similarities, existing work in natural language processing has a narrower focus as far as relational analogies between word pairs. This gap brings a natural question: \textit{can state-of-the-art large language models (LLMs) detect system analogies between narratives?} To gain insight into this question and extend word-based relational analogies to relational system analogies, we devise a comprehensive computational framework that operationalizes dominant theories of analogy, using narrative elements to create surface and system mappings. Leveraging the interplay between these mappings, we create a binary task and benchmark for Analogical Reasoning on Narratives ($ARN$), covering four categories of far (cross-domain)/near (within-domain) analogies and disanalogies. We show that while all LLMs can largely recognize near analogies, even the largest ones struggle with far analogies in a zero-shot setting, with GPT4.0 scoring below random. Guiding the models through solved examples and Chain-of-Thought reasoning enhances their analogical reasoning ability. Yet, since even in the few-shot setting, the best model only performs halfway between random and humans, $ARN$ opens exciting directions for computational analogical reasoners.
\end{abstract}

\section{Introduction}
\label{sec:introduction}

\begin{figure}[h!]
    \centering
    \includegraphics[width=1.0\linewidth]{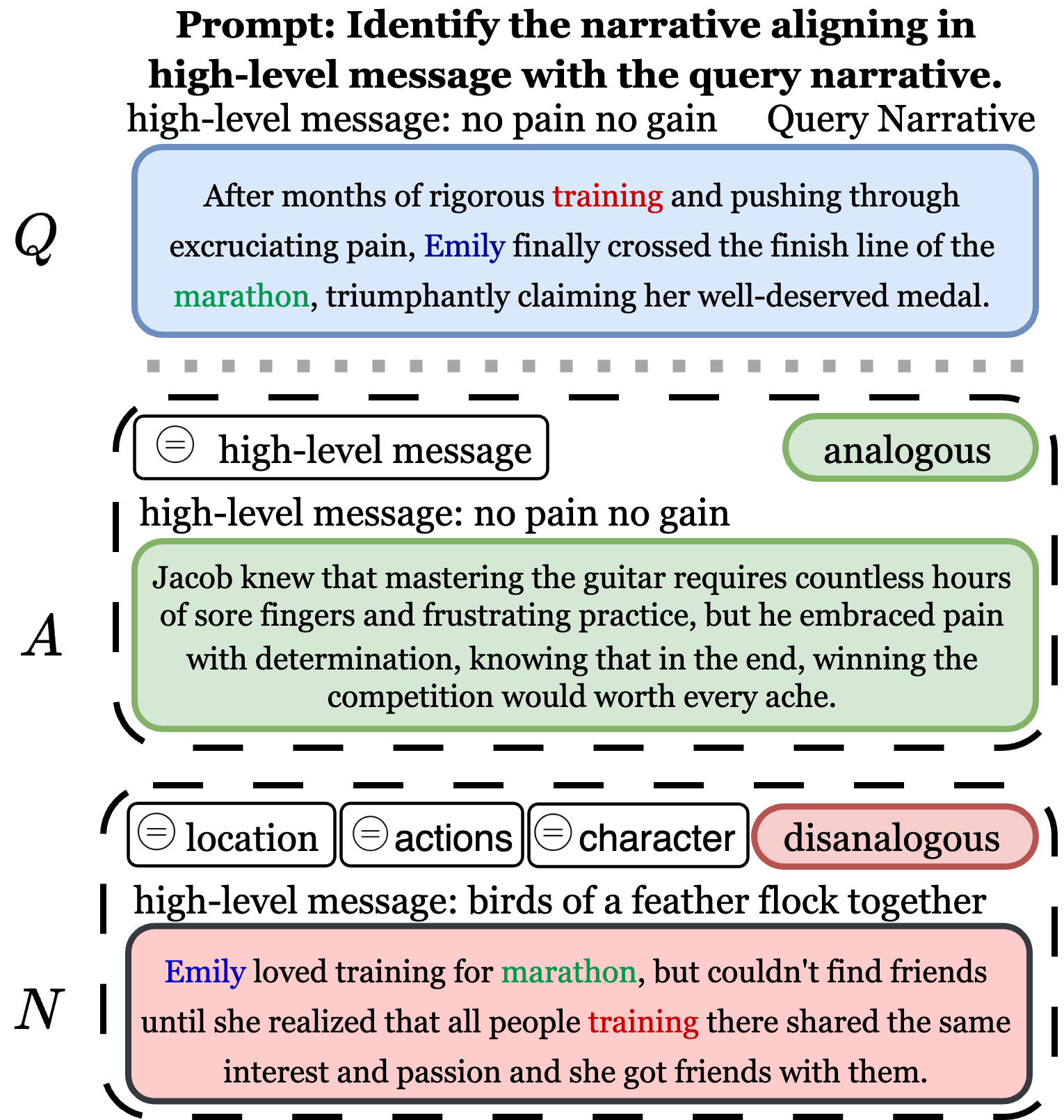}
    \vspace{-1em}
    \caption{Analogical reasoning over narratives ($ARN$): a binary task of distinguishing between analogous narrative $A$ and distractor $N$ for the query narrative $Q$. Here, $A$ represents a far analogous narrative (forming a relational system mapping) to $Q$, while $N$ is a near disanalogy (having only surface similarities).}
    \label{fig:main-fig}
\end{figure}

Analogical reasoning is a core cognitive skill unique to humans~\citep{penn2008darwin,hofstadter2001analogy}, defined as the ability to perceive and utilize the similarities between situations or events based on (systems of) relations rather than surface similarities \citep{holyoak2012analogy,gentneranalogical}.
The dichotomy between relational and surface similarity is illustrated in \autoref{fig:main-fig}: while the narratives $Q$ and $N$ overlap in terms of characters, locations, and actions (surface similarity), they fail to create a system of relational correspondences and thus are disanalogous as shown by their high-level message. Meanwhile, $Q$ and $A$ are dissimilar on the surface while forming a coherent relational system of correspondences through the high-level message: \textit{no pain, no gain}. Mappings based on systems of relations (\textbf{system mappings}) have priority over independent relations \citep{gentner1993roles}. Further, depending on whether the corresponding situations/events share the same domain or not, analogies and disanalogies can be qualified as \textit{near} or \textit{far}~\citep{gentner1983structure}.

Analogy enables creative inferences, explanations, and generalization of knowledge and has been used for scientific inventions \citep{dunbar2012scientific}, solving problems~\citep{gick1980analogical}, and policy-making \citep{houghton1998analogical}. It has also been the subject of cognitive theories and studies about humans for common processes, such as retrieval of memories \citep{WHARTON199464} and problem-solving \citep{gick1980analogical}, mostly leveraging narratives as their experimental medium~\citep[e.g.,][]{Gentner_1986,gentner1993roles,WHARTON199464}, given their multi-tiered nature and potential for abstraction. However, these studies focus on humans or models expecting well-structured inputs~\citep{lu2022probabilistic}, are small-scale \citep[Table 5]{ichien2020verbal}, and involve manually curated narratives, lacking an indication of how they can be applied to create scalable benchmarks for evaluating computational models. Nevertheless, borrowing from narratology theories~\cite{mani2013computational,Gardner2003-ym}, elements of narratives may enable analogical theories to be scaled and formalized as computational frameworks and corresponding benchmarks; however, this avenue has not been explored to date.

Meanwhile, analogical reasoning has also been relatively popular in natural language processing (NLP), typically framed as an intelligence test for models compared against humans. So-called word-based, proportional analogies of the form $(A:B::C:D)$ \citep[e.g.,][]{mikolov2013efficient,mikolov2013distributed,gladkova2016analogy,ushio-etal-2021-bert} are often used to measure the potential of word embeddings and language models in terms of analogical reasoning. Recent studies~\citep{webb2023emergent} show a strong ability of state-of-the-art (SOTA) large language models (LLMs) to discover proportional word analogies, though this skill degrades with higher complexity~\citep{wijesiriwardene-etal-2023-analogical} or associative phrasing of the input~\citep{stevenson2023large}. Shifting toward more complex settings, narrative-based analogy benchmarks that involve system mappings rather than simple word-based relational mappings, and are aligned with cognitive theories have been rarely considered~\citep{nagarajah2022understanding,wijesiriwardene-etal-2023-analogical}, with limitations in scope and generalizability. Thus, we note a significant gap between the expressivity of cognitive theories of analogies and the present analogical reasoning benchmarks in NLP. Specifically, existing computational \textit{frameworks} seldom include analogy over narratives and do not provide a formalization of key concepts such as system mappings and near/far analogs. Consequently, the \textit{benchmarks} used to evaluate SOTA models do not provide the ability for systematic exploration of the sensitivity of model performance to different analogical categories beyond word-level relations.

This gap brings a natural question: \textit{can state-of-the-art LLMs detect system analogies between narratives?} To gain insight into this question, we make three contributions: 
\begin{enumerate}
    \item \textbf{A comprehensive theory-grounded framework}, formalized in three steps: extracting narrative elements, establishing surface and system mappings that extend simple relational mappings, and inferring (dis)analogies. This framework operationalizes the link between existing analogical and narratology theories, ultimately resulting in a binary question-answering (QA) task.
    \item \textbf{A novel benchmark} for \textbf{A}nalogical \textbf{R}easoning over \textbf{N}arratives ($ARN$) containing 1.1k triples of query narratives, analogies, and distractors. Building on the underlying framework, $ARN$ contains four balanced data partitions, each characterized by the query narratives' semantic distances to its (dis)analogous narratives. 
    \item \textbf{A comprehensive study of SOTA LLMs} on the four partitions of $ARN$ in both zero- and few-shot regimes. Our experiments shed light on the ability of LLMs to distinguish surface and system similarities with and without human guidance. 
\end{enumerate}

Our experiments show a clear gap between SOTA LLMs and humans, especially in detecting far analogies.
To support further research, we make the $ARN$ benchmark publicly available at \url{https://bit.ly/3xVTjbL}.

\section{Related Work}
\label{sec:related-work}
We review existing theoretical frameworks of analogies and narrative elements that provide the foundation upon which we construct our framework. Then, we survey analogical reasoning benchmarks and compare them to ours, $ARN$.

\paragraph{Analogy frameworks.} Many categorizations exist for analogies, with a common distinction between surface and relational (system) mappings~\citep[e.g.,][]{halford1992analogical,gentner1991language,gentner1982scientific,premack1983codes}. 
\citet{gentner1983structure}'s Structural Mapping Engine (SME) defines domains and situations as systems of objects, object attributes, and relations between objects, and distinguishes between mappings formed based on each. SME defines analogies as mappings based on relations rather than objects (attributes), prioritizing interconnected systems of mappings across domains (\textbf{system mappings}). Similarly, \citet{holyoak1996mental} distinguish between three mapping categories of increasing complexity: attribute, relational, and system. Each mapping category can be formalized using a propositional representation with predicates (e.g., \textit{Mentor of}) and filler subjects (e.g., \textit{Kim, Emily}). Aligned with both frameworks, we categorize mappings into: \textbf{surface mappings} covering all categories of lower-order mappings (e.g., based on objects and object attributes), and system of relational mappings (\textbf{system mappings}) that create a coherent system of correspondences across two situations.

System mappings form analogies, having a higher priority over other mappings~\citep{gentner1983structure,holyoak1996mental}; yet, they can co-exist with surface mappings. 
Surface mappings decrease the cognitive load for recognizing system mappings since they create within-domain \textbf{(near) analogies}, also known as literal similarity~\citep{gentner1983structure}. Cross-domain \textbf{(far) analogies}, not forming surface mappings, are more challenging to identify \citep{Alexieva2017PROCESSINGDB,green2006automatic,green2008micro,green2010connecting}, promote relational thinking~\citep{vendetti2014far}, and are seen as more sound when making arguments~\citep{gentner1993roles}. Conversely, the sole existence of surface mappings yields \textbf{near disanalogies} or mere appearance, while the absence of both surface and system mappings is a \textbf{far disanalogy}, or dissimilarity~\citep{gentner1983structure}. While the distinction between near and far (dis)analogies is common in popular cognitive psychology (CogPsy) frameworks, it is unclear how to operationalize them at scale for narratives, a gap we bridge in this paper.


\paragraph{Narrative frameworks.} 
To understand what may form surface and system mappings in narratives, we review frameworks that describe narratives through their key elements.
\citet{93205451} distill four narrative elements: \textit{actors}, \textit{events}, \textit{time}, and \textit{location}. \citet{mani2013computational} distinguish between \textit{characters}, \textit{time}, and \textit{plot}. Building upon these categorizations, \citet{vossen2021narratology} expand the scope to include \textit{sequentiality} and \textit{focalizer} in addition to \textit{events} and \textit{characters}. More comprehensive and more widely adopted, \citet{Gardner2003-ym} consider \textbf{characters}, \textbf{plot}, \textbf{theme}, \textbf{setting}, \textbf{point of view}, and \textbf{style}.
To our knowledge, no prior work has leveraged such elements to infer surface and system similarities across narratives. We base our categorization of elements on categories of \citet{Gardner2003-ym}, given its comprehensiveness, to form surface and system mappings, inferring analogies and disanalogies across narratives.

\begin{figure*}[!ht]
    \centering
    \includegraphics[width=1\linewidth]{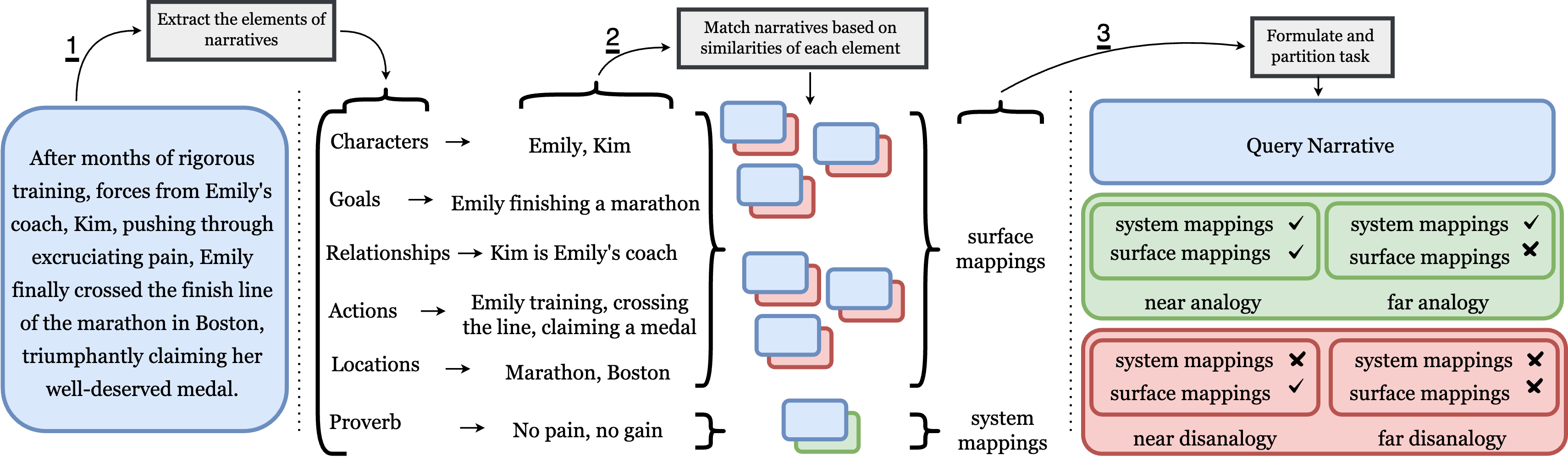}
    \vspace{-1em}
    \caption{Our proposed framework for evaluating analogical reasoning on narratives that culminates in $ARN$: 1. We start by extracting elements of narratives; 2. We then match narratives based on similarities of their extracted elements, creating corresponding mappings; 3. Based on the combinations of mentioned mappings and given the precedence of system mappings, pairs of narratives in four categories of far/near analogies and far/near distractors are organized to create the $ARN$ benchmark to evaluate LLMs' analogical reasoning in distinct scenarios.}
    \label{fig:framework}
\end{figure*}

\paragraph{Analogical reasoning benchmarks.} For both models and humans, evaluating analogical reasoning as an intelligence test \citep{hesse1965models,mitchell2021abstraction} is often posed as a word-based proportional task ($A:B::C:D$); having $Happy:Sad::C:D$, and asking to infer the missing $Angry:Calm$ or $Calm$ given $Happy:Sad::Angry:D$. Proportional analogy benchmarks~\citep[e.g.,][]{turney2003combining,jurgens-etal-2012-semeval,mikolov2013efficient,mikolov2013distributed,mikolov-etal-2013-linguistic,gao2014wordrep,gladkova2016analogy,czinczoll-etal-2022-scientific,Kotchian_Simmons_2012,Varsity_Tutors} differ in scale (up to thousands of samples), number of relations (between one and few dozens), and domains (general domain or science). Depending on the context, SOTA LLMs perform worse than humans~\citep{collier2022reality,mitchell2023comparing}, or outperform humans (which might be artifacts of training data; \citealp{lewis2024using}) in word-based analogy tasks; however, they fail to perform well in story analogies~\citep{webb2023emergent}. Notably, while cognitive theories of analogy often focus on narratives and study the distinction between surface, relational, and system similarities, existing work in NLP has a narrower focus as far as analogies between word pairs. In this study, we extend the focus from word-based relational analogies to relational system analogies in narratives. 

Meanwhile, focusing on the limited studies of matching, retrieving, or completing more complex analogies in narratives in CogPsy~\citep{ichien2020verbal}: \citet{WHARTON199464} develop 14 sets of four stories with different themes and events, used to compare human analogies between themes and events; \citet{wijesiriwardene-etal-2023-analogical} study analogies in stories using negation, entailment, and explanations, using available datasets for each; \citet{Gentner_1986} study the effect of systematicity and surface similarity of narratives' characters on analogical reasoning with 54 narratives; \citet{gick1980analogical} study analogical reasoning in problem-solving and assess the effect of similarities in problems and solutions separately. These tasks consider narrative analogies formed by humans, are thus limited in size (dozens of samples), and do not exploit the richness of the frameworks on narrative elements. Although large-scale story corpora exist~\citep[e.g.,][]{mostafazadeh2016corpus,storks2021tiered,andrus2022enhanced}, analogies are extremely scarce in free texts~\citep{sultan2022life}, and these resources are not adequate for evaluating analogical reasoning~\citep{wijesiriwardene-etal-2023-analogical} since they do not consider the theoretical dimensions of analogical reasoning from a CogPsy perspective, which is a gap we cover in this study. Most similar to our work, \citet{nagarajah2022understanding} formalize six cognitively-inspired dimensions of analogical reasoning over hundreds of Aesop's fables. However, while these dimensions can be mapped to our framework, their fine granularity limits their application to the set of fables they are devised for. Instead, we formulate a comprehensive framework that formalizes analogies and disanalogies by using narrative elements from narratology. This framework explicitly connects to the CogPsy theoretical dimensions of analogical reasoning frameworks, resulting in the inception of the first extensive, theory-grounded benchmark for analogical reasoning within narratives. Concurrently to our work, \citet{jiayang-etal-2023-storyanalogy} focus on analogies in narratives by providing a large-scale benchmark built on top of existing resources; however, they focus specifically on the distinction between surface and relational similarities. We go one step further by covering system mappings and the possible interactions between near/far analogies and distractors to paint a more accurate picture of models' analogical reasoning abilities.

\section{$ARN$ Framework}
\label{sec:extraction-framework}

Inspired by analogical reasoning frameworks in CogPsy~\citep[e.g.,][]{WHARTON199464,gentner1993roles}, we formulate \textbf{analogical reasoning over narratives} as a task where input consists of a query narrative $Q$ and two candidate narratives, $A$ and $N$ (see \autoref{fig:main-fig}). The goal is to select the narrative $A$ that is analogous to $Q$ rather than the disanalogous one $N$, based on system mappings. We frame it as a binary classification task to enable an unambiguous evaluation of the analogical reasoning abilities of models and humans \citep{frank2023baby}. We realize this task through a compositional framework with three components (\autoref{fig:framework}):
\begin{enumerate*}
    \item extraction of elements of narratives, such as characters, actions, and high-level messages (see \autoref{subsec:elements-of-narratives});
    \item formation of surface and system mappings based on those elements (see \autoref{subsec:mappings-between-narratives}); 
    \item task formulation, by defining (dis)analogies based on the mappings and partitioning the data points according to their semantic distance to the query narratives (see \autoref{subsec:analogies}). 
\end{enumerate*}

\subsection{Elements of Narratives}
\label{subsec:elements-of-narratives}
 
Following prior narratology research~\citep{Gardner2003-ym,mani2013computational,vossen2021narratology,93205451}, we consider a narrative as a piece of text that has six conceptual elements illustrated in \autoref{fig:framework}:
characters, relationships, character actions, character goals, location, and proverbial or high-level messages.\footnote{The style of narratives can, in principle, also be extracted and utilized to derive analogies, typically referred to as language style matching (LSM; \citealp{ireland2010language}). Yet, they fall outside the scope of this study, where we focus on short narratives with a uniform writing style.} To maintain generalizability in various narrative contexts, we opted for the simplest intuitive intensional definitions for each element to avoid constraining applicability and overly narrowing the scope. Moreover, both in constructing the framework and its operationalization (see \autoref{subsec:elements-extraction}), we utilized examples alongside definitions for ease of use and to mitigate confusion.

\paragraph{Characters.}
\textit{Characters} are the agents involved in narratives, including people, fictional characters, and animals. Characters can be referred to by proper names or roles, and are common elements of narratology in the literature acting as protagonists \citep{mani2013computational,vossen2021narratology,93205451}. \autoref{fig:framework} shows an example in which Emily and Kim are the characters of the narrative.

\paragraph{Relationships.}
Characters in the narrative can have different types of relationships between themselves, such as mentorship, friendship, and kinship. We consider relationships between characters to be an essential factor of the narrative \textit{plot}~\citep{mani2013computational}, as characters do not exist in isolation. An instance of a relationship between characters is Kim being Emily's mentor in \autoref{fig:framework}.

\paragraph{Character actions.} Within a narrative, character actions are pivotal as they form the \textit{plot} of the story, which unfolds through time \citep{mani2013computational}. Example actions that build up the plot of a narrative are presented in \autoref{fig:framework}: training, crossing the finish line, and claiming the medal.

\paragraph{Character goals.}
The goals of the characters drive what they do throughout the narrative. Even more, the narrative is structured around how characters approach their goals and what they do to accomplish them (known as the \textit{conflict}; \citealp{Gardner2003-ym}). An example would be Emily's goal to finish a marathon in \autoref{fig:framework}.

\paragraph{Location.}
This element refers to the \textit{setting} of the narrative, as the place or event where it is taking place \citep{93205451}. While the location is often explicitly mentioned in the narrative, it can also be implied. In the narrative presented in \autoref{fig:framework}, both the event and location are presented: marathon and Boston, respectively.

\paragraph{Proverb.}
We take the proverbial or high-level message of a narrative as an instantiation of its overarching \textit{theme} \citep{Gardner2003-ym}, which encapsulates individual elements of narratives into a coherent idea that the writer wants to convey. More formally, proverbs are short sentences containing wisdom, truth, morals, and traditional views in a metaphorical form \citep{Mieder1993}. For instance, query narrative $Q$ in \autoref{fig:main-fig} is associated with the proverb: \textit{no pain, no gain.} 

\subsection{Mappings between Narratives}
\label{subsec:mappings-between-narratives}

We leverage narrative elements to derive \textbf{surface} and \textbf{system} mappings~\citep{gentner1983structure}. We follow \citet{gentner1983structure} and \citet{holyoak1996mental} that define the order of mappings based on the level of complexity of the arguments involved.

\paragraph{Surface mappings.} 
If the mappings created between elements of two narratives are based on entities (e.g., characters), entities' attributes (e.g., characters' actions), or surface relations (e.g., the relationship between two characters), we consider them to be surface mappings, since they are based on the isolated similarity of narrative elements rather than systematic correspondences between the entirety of narratives.

\paragraph{System mappings.}
Unlike surface mappings that focus on similarities between independent elements, system mappings are formed by interconnected correspondences of relations across two narratives~\citep{gentner1983structure}. This system of correspondences can be characterized and formed by leveraging narratives' proverbs \citep{wijesiriwardene-etal-2023-analogical}; hence, two narratives with the same proverb are connected by system mappings. Similar to the moral dimensions of \citet{nagarajah2022understanding}, proverbs provide an abstract summary of the causal structure of a narrative, acting as mediators for the transition between two analogs.

\subsection{Task Definition}
\label{subsec:analogies}
\paragraph{Inference of (dis)analogies.} 

\begin{figure}
    \centering
    \includegraphics[width=0.87\linewidth]{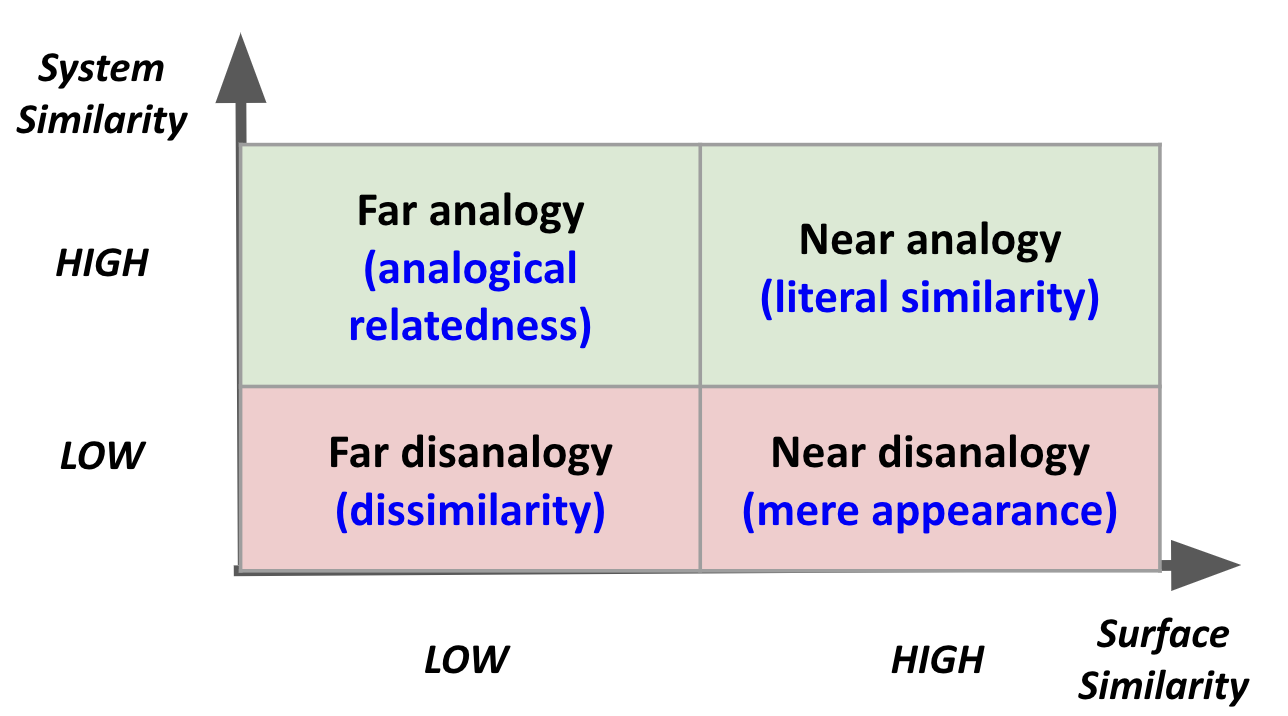}
    \caption{Interplay between surface and system similarities and analogical categories following \citet{gentner1983structure} and \citet{holyoak1996mental}.}
    \label{fig:quadrants}
\end{figure}

We consider pairs of narratives that form system mappings as analogous and the rest as disanalogous. As shown in \autoref{fig:quadrants} and similar to \citet{gentner1983structure}, we categorize analogies into far (cross-domain) analogies (analogical relatedness) that are more challenging to recognize and near (within-domain) analogies (literal similarity) that are more obvious.
Within our framework, \textbf{far analogies} are formed between narrative pairs that form system mappings without surface mappings, while \textbf{near analogies} are formed when system mappings are accompanied by surface mappings.
Surface mappings make the narratives semantically closer, while narratives that only form system mappings are semantically far from each other. 
Conversely, pairs of narratives that do not form system mappings will be regarded as disanalogies. We leverage the extent of surface mappings to distinguish near and far disanalogies (see \autoref{fig:quadrants}). Specifically, \textbf{far disanalogies} (dissimilarities) exist between narratives having none or few surface mappings, while \textbf{near disanalogies} (mere appearance) have many surface mappings.

\paragraph{Task formulation and partitioning.}

Our \textbf{binary task} is formed by coupling an analogous ($A$) and disanalogous ($N$) narrative, as candidate analogies for the same query narrative $Q$. Here, the formal distinction between near and far (dis)analogies enables a controlled setup of the relative distances between ($Q,A$) and ($Q,N$). We divide the task into \textbf{four salience-based partitions} of far/near analogies facing far/near disanalogies (see \autoref{fig:framework}), with an expectation that these partitions will affect the performance of reasoning models. For brevity, we refer to each combination by the salience of analogies and distractors, e.g., (far, near) for far analogies facing near distractors. Note that the notion of near/far analogies and near/far disanalogies, how they might face each other, and the corresponding entailments, are independent of their operationalization in different domains. We have borrowed these categorizations and distinctions broadly from analogical reasoning research in CogPsy (see \autoref{sec:related-work}), and they can be applied to any other domain too. 

\subsection{Generalizability of ARN to Narrative Analogies}

The purpose of $ARN$ is to provide a generalizable framework to study analogical reasoning in narratives grounded in narratology and analogical reasoning in CogPsy research. The provided set of elements is universal in narratives \citep{Gardner2003-ym}: entities' attributes or surface relations, proverbs as overarching messages of narratives, together with their corresponding surface and system mappings, can be broadly found in any narrative. This universality makes $ARN$ an adequate framework to describe other existing datasets. To illustrate this point, we describe how two existing benchmarks for narrative analogies can be mapped to the elements of $ARN$. StoryAnalogy~\citep{jiayang-etal-2023-storyanalogy} uses characters, actions, and relationships of narratives to represent and distinguish between entity and relational similarities while not covering higher-order mappings of analogical messages. The entity and relational elements of StoryAnalogy directly correspond to the elements in $ARN$, whereas higher-order elements and mappings, such as proverbs, are not present in this dataset. Similarly, \citet{nagarajah2022understanding} capture structural similarities between Aesop's fables concerning animal characters, actions, and morals that can be mapped to the elements in our frameworks, i.e., characters, actions, and proverbs. In terms of analogical mappings, they consider six categories of Shallow and Deep Attribute, Relational, Event, Structural, and Moral analogies. Moral analogies fall under system mappings in our framework, while others, being more focused on the surface similarities, can be mapped to specific mappings that we define based on characters, relationships, actions, goals, and locations — e.g., Event analogies to surface mappings based on actions — all of which fall under surface mappings. While certain scenarios might require further adaptations in the operationalization of the framework, we expect that the three-tier structure of the $ARN$ framework to be used as the basis for both existing and future tasks, as the underlying concepts of surface/system mappings, far/near (dis)analogies, and the four partitions covering the interplay between them are well grounded in theory and generalizable across domains and contexts.

\section{$ARN$ Benchmark}
\label{sec:realization-of-the-framework}

In this section, we describe the realization of the framework in \autoref{sec:extraction-framework} as a benchmark for \textbf{Analogical Reasoning over Narratives ($ARN$)}. 
Without loss of generality, we use narratives from the ePiC crowdsourced dataset \citep{ghosh2021epic} that consists of 2,500 short narratives (on average 4-8 sentences) annotated with 250 corresponding proverbs (10 narratives per proverb).
We chose the ePiC dataset because of its systematic annotation of proverbs and the wide range of topics that it covers, allowing for the extraction of diverse elements with corresponding mappings. 

In \autoref{subsec:elements-extraction}, we explain how we extract elements from the narratives in ePiC. The mappings between these elements are formed by quantifying their similarity across pairs of narratives (see \autoref{subsec:mappings-extraction}). The formed surface mappings serve as a basis to find distractors (see \autoref{subsec:selecting-distractors}), whereas the system mappings form corresponding analogies (see \autoref{subsec:selecting-analogies}).
The $ARN$ benchmark consists of ($Q,A,N$) triples, pairing analogies $A$ and their corresponding distractors $N$ for the same query narrative $Q$. The semantic similarity between $Q$ and $A$, and between $Q$ and $N$, are defined by using the number of formed surface mappings. The statistics and quality analysis of $ARN$ are presented in \autoref{subsec:dataset-statistics-quality}. For more details about the involved preprocessing steps and examples, refer to \autoref{appendix:data-collection-details}.

\begin{table}[!t]
\centering
\scriptsize
\begin{tabular}{p{1.4cm}|p{5.0cm}}
\toprule
Element & Examples \\
\midrule
Characters & teacher, student, speaker, Kelly, Kim, John, driver, judge, painter, wife, neighbor \\
\midrule
Relations & friend, sibling, seller-buyer, manager-employee, roommate, classmate, coworker \\
\midrule
Actions & dating, eating food, borrowing money, attacking enemy, rejecting tasks, being happy \\
\midrule
Goals & to borrow money without paying it back, be prepared for unexpected problems during cross-country driving trips, to have more rest and be happier \\
\midrule
Locations & work, river, graduation, home, garden, bed, hospital, company, wedding, dealership\\
\midrule
Proverbs & There's no accounting for tastes, You are never too old to learn, It's no use crying over spilt milk\\
\bottomrule
\end{tabular}
\caption{
Examples of narratives' elements across the $ARN$ benchmark.
}
\label{tab:elements-examples}
\end{table}

\subsection{Extracting Narrative Elements}
\label{subsec:elements-extraction}
A key benefit of the ePiC dataset is the already annotated proverbs associated with each narrative.\footnote{We do not use span annotations in the ePiC dataset.} \autoref{tab:proverbs-narratives} in the Appendix provides examples of narratives alongside their proverbs. We extract the other elements of a narrative using an LLM (GPT3.5; \citealp{chatgpt}) denoted by $llm(c, n)$ for extracting element $c$ from a narrative $n$. More concretely, we explicitly prompt the LLM, asking it to extract the element, $c$, from the provided narrative, $n$, returning a Python List object. We opted to utilize GPT3.5 for this task to save both time and money, given its high accuracy in simple information extraction tasks \citep{wei2023zeroshot}. Sample extracted elements from various narratives are shown in \autoref{tab:elements-examples}, which demonstrate the diversity of the subjects covered in the ePiC dataset. Across all narratives, on average, we extracted 4.55 actions ($SD = 1.90$), 2.44 goals ($SD =  1.46$), 2.59 characters ($SD = 0.96$), 1.71 relations ($SD = 1.16$), and 3.90 locations ($SD =  3.67$). After evaluating the extracted elements by a graduate student who was instructed to check whether the extracted elements were present in a subset of 100 random narratives, the precision of above $94\%$ was observed across all elements. We focus on precision in this pilot study since created mappings between two narratives based on a certain element do not require identical elements. Additional details are provided in \autoref{appendix:extracting-narratives-elemenets}.

\subsection{Forming Mappings}
\label{subsec:mappings-extraction}
We approach forming system mappings and surface mappings differently. As the proverbs in the ePiC dataset form a closed set, where each narrative is associated with exactly one proverb, we form system mappings between narratives that have (nearly) identical proverbs through automatic matching. The remaining five elements may consist of multiple values (e.g., multiple actions in a plot), each with an arbitrary phrasing. Therefore, the similarity between those elements forms a continuum rather than a dichotomy which was the case with proverbs. Inspired by the context-aware characteristics of Transformers~\citep{vaswani2017attention} rendering them strong at capturing surface similarities in isolation (e.g., between actions of two narratives), we estimate the semantic similarity between the values for element $c$ in two narratives $n_1$ and $n_2$ with the cosine similarity of sentence BERT (all-mpnet-base-v2; \citealp{reimers-2019-sentence-bert}) encoding of those values:

\begin{fleqn}[0pt]
\setlength{\abovedisplayskip}{6pt}
\setlength{\belowdisplayskip}{6pt}
\begin{alignat*}{2}
sim(n_1, n_2, c) = \cos(&sbert(llm(c, n_1)), \\
&sbert(llm(c, n_2)))
\end{alignat*}
\end{fleqn}
Here, the element $c$ comes from the set $C = \{characters, relationships, goals, locations, \\actions\}$. A surface mapping between $n_1$ and $n_2$ for element $c$ is formed when $sim(n_1, n_2, c)\geq t$, where $t$ is a manually defined threshold. Note that the short narratives we focus on in these studies allow us to look at mappings based on each narrative element broadly and report whether a mapping exists or not. However, in long narratives, it might be necessary to focus on more fine-grained associations and per element, distinguish between different element items and their corresponding mappings across narratives.

\subsection{Selecting Distractors}
\label{subsec:selecting-distractors}
To gather distractors, first, all pairs of narratives that create surface mappings as extracted in \autoref{subsec:mappings-extraction} are retrieved, and the pairs with the same proverbial messages are removed to ensure that the pairs are disanalogous. The retrieved pairs are categorized into far and near disanalogies based on the number of surface mapping types they contain (we consider narratives that have at least three surface mappings as near and less than three as far). This step yields $m = 548$ pairs of query narratives and distractors in the $ARN$ dataset: $P=\{(Q_1, N_1), (Q_2, N_2), \dots, (Q_m, N_m)\}$.

\subsection{Selecting \& Generating Analogies}
\label{subsec:selecting-analogies}
The narratives sharing the same proverb in the ePiC dataset are semantically far from each other by design. We, thus, leverage this property to establish far analogies between pairs of narratives that share the same proverb. To guarantee low surface similarity, for each query narrative $Q_i$ in $P$, we select the narrative with the same proverb that is semantically the farthest from $Q_i$ (using sentence BERT; \citealp{reimers-2019-sentence-bert}; and cosine similarity) as a far analogy.

Near analogies are formed between pairs of narratives that form both surface and system mappings, and narratives associated with the same proverb in ePiC do not meet this requirement. To generate near analogies for a query narrative $Q_i$ in $P$, we use a hybrid approach, utilizing both the abilities of LLMs (GPT3.5) and humans. First, we prompt GPT3.5 to generate seed narratives, given restrictions on the proverbial message that the generated narrative should reflect and the narrative elements it should mention. Then, upon an observation that GPT3.5 does not generate narratives with given proverbs consistently, authors inspected the generated narratives and edited their final versions manually to ensure high quality. More details are provided in \autoref{appendix:generating-near-analogies}.

\subsection{Benchmark Statistics \& Quality}
\label{subsec:dataset-statistics-quality}
The final $ARN$ benchmark consists of $1096$ triples of query narratives, distractors, and analogies. $ARN$ has 294 far disanalogy pairs and 254 near disanalogy pairs, each creating triples with both a far analogous narrative and a near analogous narrative to the query narratives, which makes the total size of the dataset balanced equally between far and near analogies.

All the triples in the dataset were manually investigated by the first and the last author to remove or revise the ones with possible ambiguities. There were less than 30 triples that needed revisions or were completely removed. Further, to ensure the fairness of the proposed task and the quality of the $ARN$ dataset as a realization of this task, we ran experiments to measure human performance. Two research assistants were instructed (find instructions at \url{https://bit.ly/48QFUyA}) to identify the analogous narratives in $ARN$ for 120 data points (over $10\%$) of the dataset consisting of 30 samples for each combination of far/near analogies and far/near distractors. The mean accuracy of the participants in the last row of \autoref{tab:main-results} shows a high consensus between the perceived system mappings of the participants and our benchmark. Moreover, the Cohen Kappa IAA \citep{mchugh2012interrater} score between the annotators on this task was $0.865$, demonstrating high agreement. By a manual inspection, disagreements were mostly caused by wrong predictions by one of the participants. We did not see any trends between disagreements and data attributes (e.g., whether it is the far analogies or near analogies causing disagreements).

\section{Experiments}
\label{sec:results}
We evaluate the analogical reasoning abilities of language models over narratives in $ARN$ using two settings.\footnote{We did not find a difference between the models' performance on all data points and data points where annotators had an agreement on.} First, we report on their \textit{autonomous} \citep{Radford2019LanguageMA} analogical reasoning abilities to assess the emergent abilities of LLMs \citep{wei2022emergent} and the latent analogical reasoning abilities they might have obtained during their training (see \autoref{subsec:results-autonomous-analogical-reasoning}).
Following a zero-shot formulation similar to \autoref{fig:main-fig}, we provide the model with the query narrative and the two candidate narratives, asking for the best analogical match to the query narrative. Asking for the most \textit{similar} narrative yielded similar results (see \autoref{appendix:do-models-find-analogical-relatedness-and-similarities-different}). We evaluate the autonomous analogical reasoning abilities of six LLMs: GPT3.5 \citep{chatgpt}, GPT4.0 \citep{openai2023gpt}, UnifiedQA \citep{khashabi2020unifiedqa}, Llama-2 \citep{touvron2023llama}, FlanT5 \citep{chung2022scaling}, and Macaw \citep{tafjord2021general} using deterministic hyperparameters ($Temperature = 0$). We also evaluate sentence BERT (SBERT, all-mpnet-base-v2; \citealp{reimers-2019-sentence-bert}) by picking the narrative with a higher cosine similarity to the query narrative as a baseline.
Additionally, in a more \textit{guided} scenario, we experiment with a few-shot (in-context learning; \citealp{brown2020language}) setting to enhance models' abilities with solved demonstrations (see \autoref{subsec:results-guided-analogical-reasoning}). We evaluate Llama-2, GPT3.5, and GPT4.0 because of their in-context learning abilities and long context windows. Solved demonstrations are chosen randomly, and performance is reported as the average on three runs. For more details about the experimental setup and prompts, refer to \autoref{appendix:experimental-details} and \autoref{appendix:prompts}, respectively.

\subsection{Autonomous Analogical Reasoning}
\label{subsec:results-autonomous-analogical-reasoning}

\begin{table}[!t]
\centering
\small
\resizebox{\linewidth}{!}{%
\begin{tabular}{p{2.3cm}|c|c|c|c|c}
\toprule
\thead{Task Partition} & \thead{(near, far)} & \thead{(near, near)} & \thead{(far, far)} & \thead{(far, near)} & \thead{$Avg.$} \\
\midrule
SBERT & 84.3 & 71.5 & 12.6 & 1.00 & 42.3 \\
GPT3.5 & 88.1 & 81.3 & 50.4 & 21.7 & 60.3 \\ 
GPT4.0 & \textbf{94.0} & \textbf{92.5} & \textbf{57.1} & 29.1 & \textbf{68.1} \\ 
UnifiedQA-L & 43.2 & 49.1 & 47.2 & 50.5 & 47.5 \\
UnifiedQA-3B & 66.4 & 68.4 & 47.3 & 44.4 & 56.6\\ 
UnifiedQA-11B  &  60.7 & 61.2 & 54.8 & \textbf{74.6} & 62.8 \\ 
Llama-2-7B & 63.4 & 58.0 & 50.1 & 43.1 & 53.7 \\ 
Llama-2-13B & 80.9 & 81.4 & 44.5 & 35.4 & 60.5\\ 
FlanT5-L & 84.5 & 80.3 & 41.4 & 14.4 & 55.1 \\ 
FlanT5-xl & 78.9 & 68.3 & 44.7 & 21.1 & 53.2 \\ 
FlanT5-xxl & 89.9 & 81.3 & 51.1 & 35.6 & 64.4 \\
Macaw-11B & 88.0 & 84.6 & 42.1 & 35.8 & 62.6 \\
\midrule
$Avg.$ & 76.9 & 73.2 & 45.3 & 33.9 & 57.3 \\
\midrule
human & 98.6 & 97.2 & 96.8 & 91.4 & 96.0\\ 
\bottomrule
\end{tabular}
}
\caption{
Accuracy (with a random baseline of $50\%$) of tested LLMs in \textbf{zero-shot setting}, on $ARN$ in four categories of far/near analogies facing far/near disanalogies, e.g., (near, far) indicates the setting where near analogies are alongside far disanalogies. \textbf{Boldfaced} numbers in each partition show the best-performing model in that partition.
}
\label{tab:main-results}
\end{table}

\paragraph{How does analogical reasoning of models compare to humans?}
In \autoref{tab:main-results}, we observed that models are not as good as humans at distinguishing analogies from distractors ($57$ vs $96\%$). On near analogies, the humans' average performance was $97.9\%$, whereas the models' performance averaged $75.0\%$, which is halfway between random and human performance. Focusing merely on semantic similarity (using SBERT) already yielded a good performance in this setting ($77.9\%$) which confirms the general similarity understanding of language models. Intuitively, the human-LLM gap was wider in solving far analogies where analogies have little surface similarity to query narratives. Although far analogies were slightly more difficult for humans, as evident in the small drop in performance, the drop was significant in LLMs. While humans' average performance was $94.1\%$, models only achieved $39.6\%$ on average (worse than random performance) when solving far analogies. Similarly, we also observed that in both near and far analogies, as distractors get semantically more similar to the query narratives (far $\rightarrow$ near distractor), both LLMs' and humans' performance drop, with LLMs experiencing more notable drops. The most competitive model, GPT4.0, performed close to humans on near analogies; however, its performance was below random in far analogies. The relatively low performance of SOTA models underlines the challenging nature of analogical reasoning, even for the most capable LLMs.

\paragraph{How does the relative distance between the analogies and distractors affect models' performance?}
Comparing the performance of the models in different partitions of $ARN$, we observed the following patterns. All models performed well in choosing the near analogy over the far disanalogy ($76.9\%$), i.e., cases where analogies form surface mappings on top of system mappings and distractors do not form any mappings with the query narratives. Facing two candidates with similar surface similarity (near-near or far-far), models' performance dropped significantly. Models scored $73.2\%$ on average in a (near, near) setting which shows that models recognize system mappings in the presence of surface mappings, even though the distractor also forms surface mappings with the query narrative. Models performed much worse in the (far, far) setting (scoring only $45.3\%$), which shows their disability to recognize system mappings in the absence of surface mappings. This observation was further emphasized by the most challenging partition, where models needed to prioritize far analogies over near distractors, as models failed to prioritize system mappings over surface mappings (scoring $33.9\%$), regardless of their prompt instructions specifically asking for system mappings. Note that although solving examples in all partitions of the data requires analogical reasoning, the reasoning pattern specifically involved in each partition can be substituted by certain heuristics (e.g., the reasoner can achieve great scores by choosing the dissimilar narrative in the [far, near] condition without involving analogical reasoning). We run our experiments completely isolated from each other, which makes the active development of such heuristics highly unlikely. However, it would be an interesting research direction to study the heuristics that models might use in analogical reasoning.

\paragraph{How do models' analogical reasoning abilities compare to one another?}
As expected, the best-performing model was GPT4.0, with an average performance of $68.1\%$, and the worst-performing model was SBERT, with an average performance of $42.3\%$. GPT4's performance was in line with prior findings about analogical reasoning being one of the emergent abilities of LLMs \citep{webb2023emergent,wei2022emergent}; however, its performance on $ARN$ was still halfway between random and human performance. Meanwhile, the poor performance of SBERT was expected since it relies on surface semantic similarity between narratives, which makes it perform relatively well on recognizing near analogies but reaches only $1\%$ when prioritizing far analogies. While generally, we saw that larger models perform better in the first three partitions, we observed that the best-performing model in the (far, near) setting is UnifiedQA-11B, even outperforming the GPT models with orders of magnitude more parameters. We attribute this curious observation to the smaller and more structured (instruction) fine-tuning that UnifiedQA has undergone and its alignment with the binary question-answering (QA) format of our task. To investigate this observation further, we also evaluated other QA models: FlanT5, Macaw, and smaller UnifiedQA variants. Aligned with our hypothesis, We observed that the other QA models also perform relatively well, scoring on par or better than the much larger foundational models: GPT3.5, GPT4.0, and Llama-2, on the (far, near) partition.

\begin{figure}[h]
    \centering
    \includegraphics[width=1.0\linewidth]{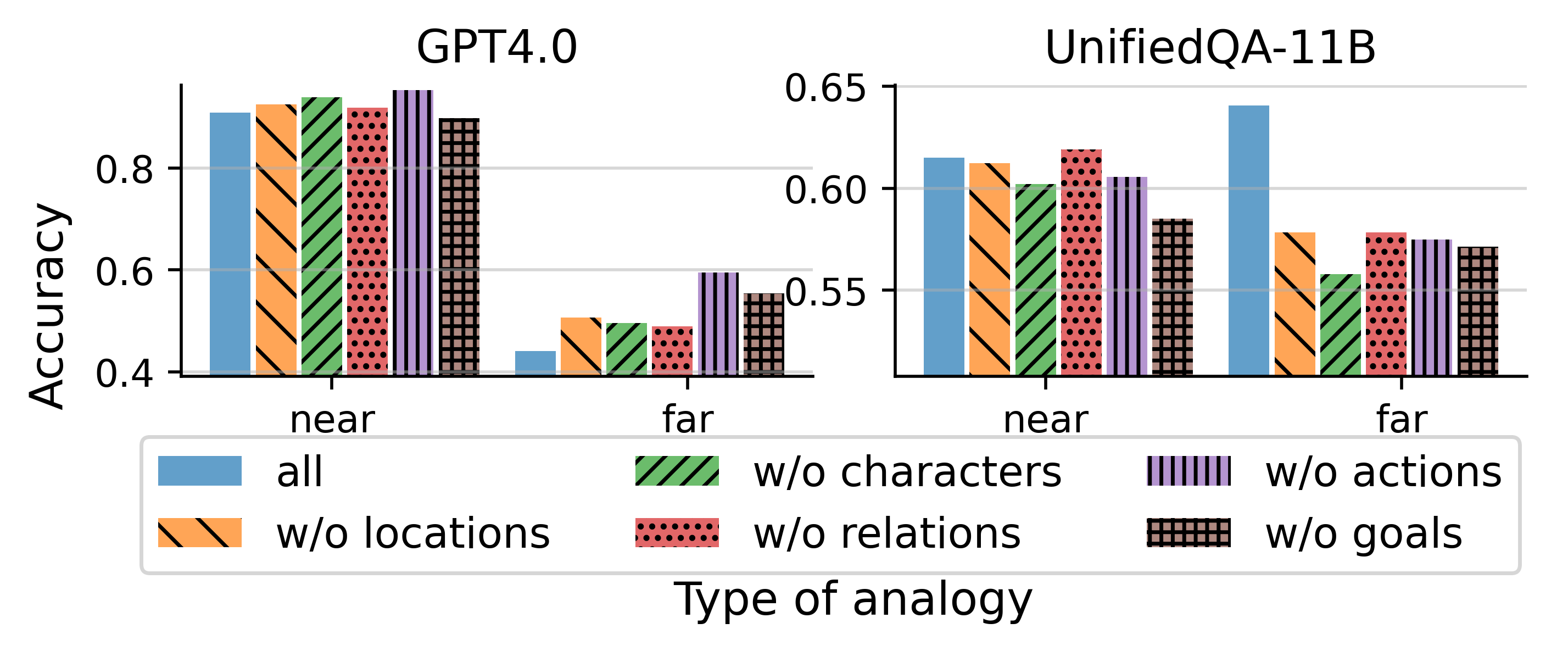}
    \vspace{-1.4em}
    \caption{GPT4.0 and UnifiedQA-11B's performance grouped jointly by the type of analogies and distractors.}
    \label{fig:effect-of-contextual-dimensions-distractors}
\end{figure}

\paragraph{Which surface mappings are more distracting for models?}

We further studied the distracting effect of each narrative element on the analogical reasoning abilities of the two best-performing models, GPT4.0 and UnifiedQA-11B. \autoref{fig:effect-of-contextual-dimensions-distractors} demonstrates the performance of models when distractors from each narrative element were excluded in the experiments, which we consider to be a proxy for the distraction effect of each narrative element — if excluding narrative elements $c_1$ from the distractor sampling leads to higher performance compared to excluding $c_2$, we infer that that $c_1$ has a higher distracting effect compared to $c_2$. For GPT4.0 (and all other models except UnifiedQA; see trends for other models in \autoref{appendix:Distracting-Effect-of-Surface-Mappings}), the aggregated distracting effect of all elements was stronger than individual elements, demonstrating the distinct effect of each element. Among the different elements, mappings based on actions led to the most effective distractors both when detecting far and near analogies. In contrast, for UnifiedQA-11B, the aggregated effect of all surface mappings was not the most distracting, especially in far analogies where the model had the highest performance when all types of distractors were included. This observation, alongside the model having the highest performance in far analogies across all models, shows the model's higher robustness for analogical reasoning, where it may ignore and even benefit from distracting surface mappings formed with disanalogies. 

\paragraph{Error analysis.}
Based on our observations on the free-text rationales and predictions by GPT4.0, the explanations generated by models were indeed based on high-level messages. However, these explanations could be incorrect in different ways that we iterate on some of the more salient ones:
\begin{itemize}[label={$\bullet$}, topsep=5pt, itemsep=2pt, leftmargin=10pt]
    \item The rationale was correct, while the prediction was incorrect. 
    In the first example in \autoref{tab:errors} in the Appendix, the explanation was aligned with the correct choice, but the model predicted the second narrative by mistake.
    \item The provided explanation followed the main theme of the narratives and was aligned with the keywords of the query narrative and the chosen narrative. However, the explanation failed to focus on the high-level message of the query narrative. In the second example shown in \autoref{tab:errors} in the Appendix, the prediction was made based on narratives' keywords. However, the high-level message that the correct answer is based on was not recognized by the model.
    \item The rationale captured the high-level message of the query narrative and the chosen narrative but disregarded the connection between the two narratives, i.e., the fact that a mapping should be formed between narratives. In the last example in \autoref{tab:errors} in the Appendix, the explanation and the gathered high-level message were aligned with the chosen narrative but did not hold true in the query narrative.
\end{itemize}

\subsection{Guided Analogical Reasoning}
\label{subsec:results-guided-analogical-reasoning}

\begin{figure}[h]
    \centering
    \includegraphics[width=1.0\linewidth]{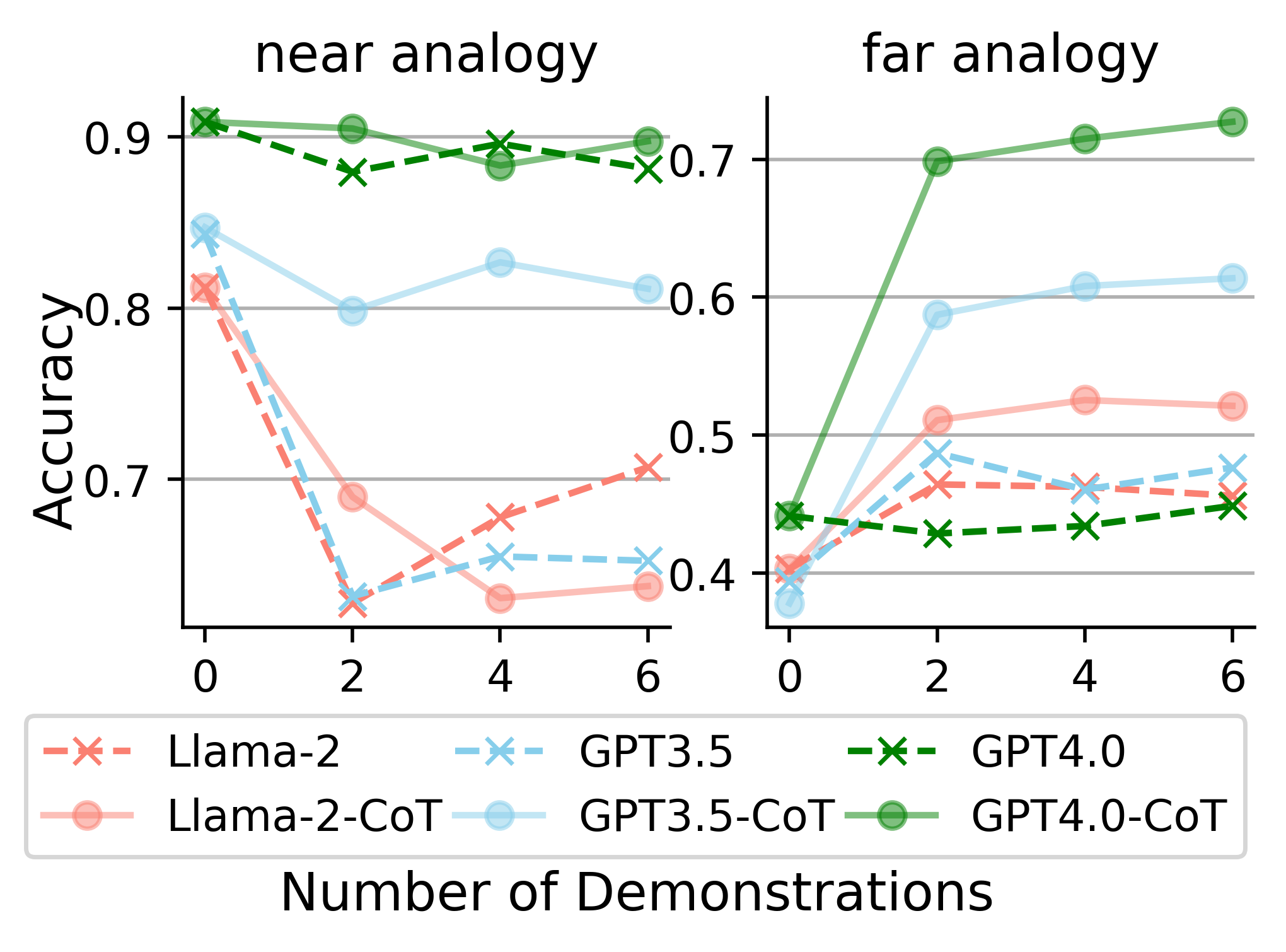}
    \vspace{-1.4em}
    \caption{Performance of Llama-2, GPT3.5, and GPT4.0 on $ARN$ where $\{0, 2, 4, 6\}$ randomly chosen solved demonstrations from both far and near analogies were shown to the model as hints. Demonstrations were provided both as normally solved demonstrations and Chain-of-Thought reasoning, denoted as CoT.}
    \label{fig:in-context-learning}
\end{figure}

\paragraph{Does providing solved demonstrations improve analogical reasoning?}
A natural follow-up question is whether the performance of LLMs can be enhanced by guiding them through solved demonstrations in comparison to a zero-shot setting.
The performance of different LLMs in a few-shot setting is demonstrated in \autoref{fig:in-context-learning} denoted by $\times$ signs, alongside their performance in a zero-shot setting where the number of demonstrations is set to zero. 
When models were provided with solved demonstrations, their near-analogy skills degraded, whereas their far-analogy performance increased compared to the zero-shot setting. GPT4.0 was the most stable model, with its performance slightly increasing in far analogies (especially with more demonstrations), with a small drop in performance in near analogies. The other models, GPT3.5 and Llama-2, had more emphasized trends: they obtained a much larger gain on far analogies (close to 10 absolute points) while also losing much of their performance on near analogies, though more demonstrations mitigated part of the loss. Overall, these results suggest that the models do not efficiently leverage the solved demonstrations, and surprisingly, their performance in a few-shot setting is not necessarily better than zero-shot setting. We leave a deeper investigation of the models' behavior to future work as they are beyond the scope of this paper.

\begin{figure}[h]
    \centering
    \includegraphics[width=1.0\linewidth]{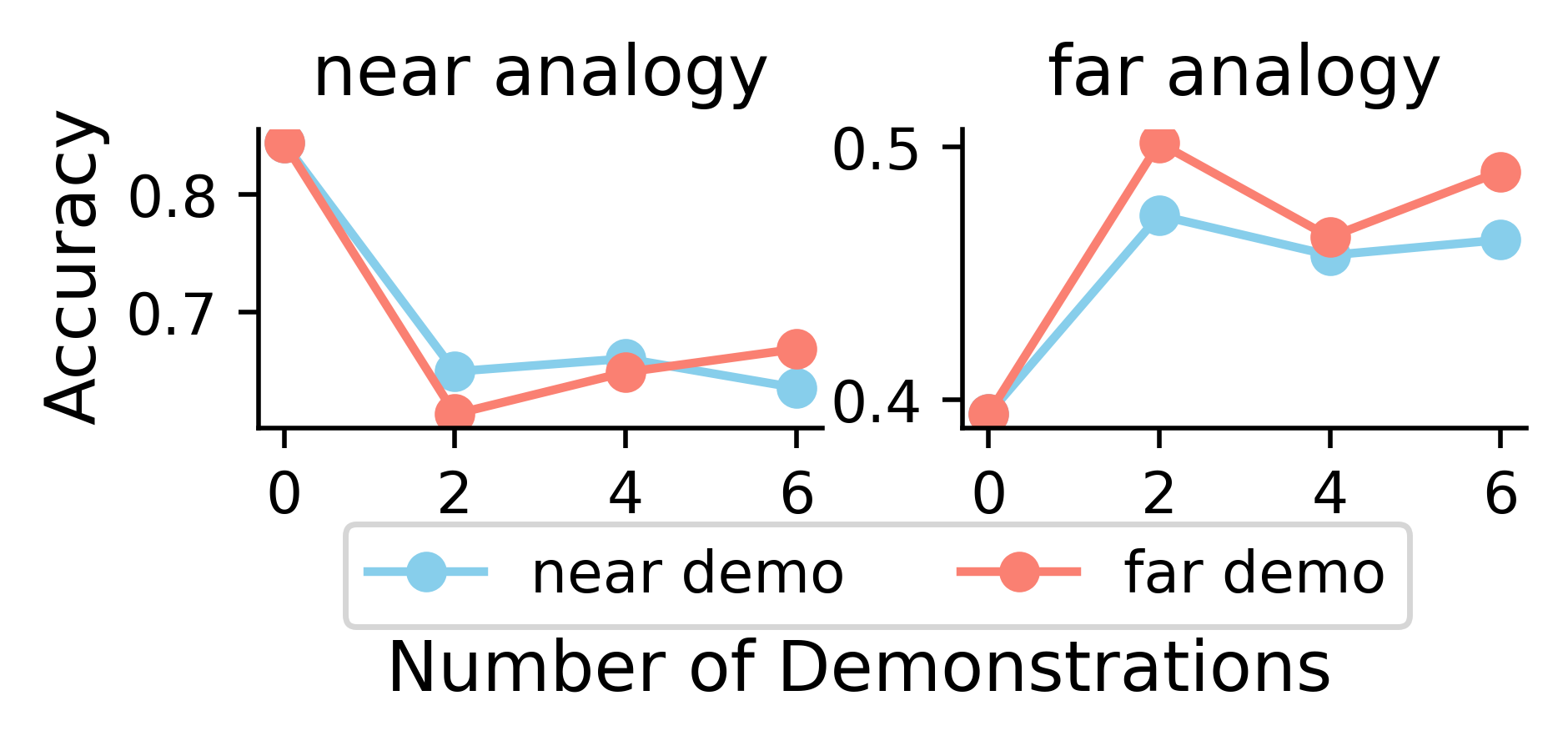}
    \vspace{-1.4em}
    \caption{Performance of GPT3.5 on $ARN$ where $\{0, 2, 4, 6\}$ solved demonstrations from either far or near analogies were shown to model to serve as hints.}
    \label{fig:in-context-learning-per-demo-chat-gpt}
\end{figure}

\paragraph{How does the type of knowledge in solved demonstrations influence their effectiveness?}
The models may use solved demonstrations that cover near or far analogies differentially, which prompted us to test including either far or near analogies in the provided demonstrations to LLMs. Due to limitations in the expenses of using OpenAI's API for GPT4.0, we conducted this experiment on GPT3.5 and Llama-2.
The result of these experiments for GPT3.5 is presented in \autoref{fig:in-context-learning-per-demo-chat-gpt} (Llama-2 showed similar trends). Overall, we observed similar patterns when using either near or far analogy demonstrations. Using either near or far analogy demonstrations improved the performance of models on detecting far analogies, while both demonstration types were detrimental to the models' performance facing near analogies. However, increasing the number of far analogy demonstrations gradually improved the detection of near analogies, which suggests that far analogies can be slightly more helpful as demonstrations. The advantage of using far analogies as demonstrations can be explained by them being independent of the domains or specific content of analogs \citep{holyoak1996mental}.

\paragraph{How does step-by-step reasoning affect the performance of the models?}
Solved demonstrations of analogies hold a promise to enhance performance while simultaneously being challenging to process, which may have led to the counter effects on model performance in \autoref{fig:in-context-learning}. Inspired by these observations, we investigated whether guiding the model to reason step-by-step can enable it to perform better. We employed Chain-of-Thought (CoT) prompting \citep{wei2023chainofthought}, given its positive impact on tasks that require complex reasoning. We experimented with Llama-2, GPT3.5, and GPT4.0 using $\{2, 4, 6\}$ demonstrations equally split between far and near analogies. In this step-by-step procedure, the model was instructed to extract the proverbial message of each narrative, followed by generating the final predictions. In general, CoT demonstrations helped the models more than normal demonstrations (see \autoref{fig:in-context-learning}). In near analogies, CoT demonstrations still hurt the performance compared to the zero-shot setting, although this negative effect was less strong than in the normal few-shot setting. However, in far analogies, CoT demonstrations were more beneficial than both the normal few-shot demonstrations and no demonstrations (zero-shot setting). Here, GPT-based models gained up to 30 absolute points compared to the zero-shot setting. The models' performance not improving for near analogies is likely because the models are already using surface similarity cues to solve the task, and CoT's reasoning chains are not recognized as critical by the models. The improvements in far analogies are likely because models need to look for elaborate mappings without surface similarity, and CoT can facilitate this by breaking the task into smaller, more manageable parts.

\section{Conclusions and Future Work}
This paper addressed a gap between analogical reasoning theories in cognitive psychology and analogical reasoning benchmarks in NLP. Our proposed framework studies analogical reasoning over narratives by a three-tier process of extracting narrative elements following narratology, devising corresponding mappings according to analogical reasoning theories, and providing a formal method for quantifying near/far analogical distances to create near/far analogies and disanalogies. The framework provides a solid basis for a binary QA task that enables a systematic study of models' analogical reasoning abilities across various scenarios. Leveraging data on narratives with high-level proverbial messages, we instantiate the devised task with the Analogical Reasoning on Narratives ($ARN$) benchmark, consisting of 1096 triples of query narratives, analogies, and distractors. Evaluating multiple LLMs on $ARN$ in a zero-shot setting suggests that while models can recognize near analogies, their analogical reasoning performance degrades when detecting far analogies, characterized by the absence of surface mappings. This trend also holds for GPT4.0, performing best on average but dropping to a below-random performance on detecting far analogies in a zero-shot setting. Few-shot prompting with Chain-of-Thought reasoning enhances models' performance in far analogies while being detrimental to solving near analogies.
Overall, we show that LLMs' analogical reasoning over narratives lags behind humans, especially on far analogies, which motivates further research on devising computational analogical reasoners on narratives.

\section{Limitations}
While our framework showcases its potential within a specific narrative benchmark, extending its application to diverse benchmarks and resources might require additional adaptations. We see additional development of benchmarks and discussions around the analogical reasoning capabilities of LLMs extremely useful as they promote generalizability and reusability of models in arbitrary domains. Further, although we used narrative elements to form mappings, alternative approaches like using elements inspired by problem-solving can be explored~\citep{gick1980analogical}. Finally, going beyond LLMs, considering other methods inspired by CogPsy and neuro-symbolic models might offer valuable insights into a more diverse set of models and their analogical reasoning on narratives.

\section*{Acknowledgements}
This research was supported, in part, by the Army Research Laboratory under contract W911NF-23-2-0183, and by National Science Foundation under Contract No. IIS-2153546. We thank Keith Holyoak and Hongjing Lu for the fruitful discussion around the topic in the initial stages of the study. Moreover, we thank the reviewers and action editors from TACL for their insightful comments and valuable feedback.

\bibliography{tacl2021}
\bibliographystyle{acl_natbib}


\appendix

\section{Data Collection Details}
\label{appendix:data-collection-details}

As we use the ePiC dataset \citep{ghosh2021epic} as our narrative database, there are certain pre-processings that needed to be done on the original ePiC dataset before they would be ready for the subsequent steps explained in \autoref{sec:realization-of-the-framework}. We modify all the narratives that do not necessarily reflect the proverb they are supposed to be associated with by rewriting the narratives and keeping the elements of the narratives intact.\footnote{This happens since all the narratives are crowdsourced, and limited supervision or structure has been in place when gathering the narratives.} We consider two proverbs with the same meaning similar, and hence, the narratives associated with those proverbs are, in turn, analogous. On this ground, we consolidate proverbs and their corresponding narratives that convey the same meaning into a single representative proverb with its associated narratives, which results in the cleaned dataset having 223 unique proverbs and between 10 to 40 narratives associated with each proverb.

In the process of forming the mappings between lower-order elements, many similarities exist between the 2500 narratives we have in the original dataset. We sort the similarities and only keep the most meaningful ones by a manually set threshold starting from the top. When focusing on the similarities based on a single element $c$, we do this sorting based on $sim(n_1, n_2, c)$ and $sbert$ similarity. However, when focusing on the similarities based on multiple elements where multiple mappings are formed simultaneously, we sort by the aggregated similarity of two narratives based on elements $C' \subset C$ computed by $\sum_{c \in C'} log(sim(n_1, n_2, c))$.

\begin{table*}[!ht]
\centering
\small
\begin{tabular}{p{0.17\linewidth} | p{0.73\linewidth}}
\toprule
\textbf{Proverb}  &  \textbf{Narrative} \\
\midrule
Practice what you preach & Tommy grabbed a cookie and began to eat it. "Don't do that!" screamed Tommy's dad, "You will spoil your dinner if you eat it now." Tommy puts the cookie down. Soon after, Tommy's dad picks up the cookie and finishes it. Tommy walks in a question why his dad is eating the cookie if it will spoil dinner. \\
\midrule
Chain is only as strong as its weakest link & The ingenious security system overlooked the possibility of an unscrupulous IT manager which was the only possible way to hack system. The system was hacked weeks later and the company lost millions. \\
\midrule
A miss is as good as a mile & Kate nervously awaits her test results. This is her last chance to pass the SAT before college. As she looks at her paper, she realizes that she once again did not get what she needed to get into her school of choice. Her score was better than last time, but nonetheless, she still had failed. \\
\bottomrule
\end{tabular}
\caption{
Examples of narratives with their associated proverbs.
}
\label{tab:proverbs-narratives}
\end{table*}

\autoref{tab:proverbs-narratives} provides examples of narratives alongside their high-level proverbial messages. Further, \autoref{tab:arn-examples} demonstrates examples of the $ARN$ dataset covering four partitions mentioned in \autoref{subsec:analogies}.

\begin{table*}[ht]
\centering
\small
\scalebox{0.8}{
\begin{tabular}{p{0.10\textwidth}| p{0.33\textwidth} | p{0.32\textwidth} | p{0.34\textwidth}}
\toprule
\textbf{Partition}  &  \textbf{Query narrative} & \textbf{narrative 1} & \textbf{narrative 2} \\
\midrule
far, near & Johnny was having a very hard time at work. He had too many projects and too many short deadlines, and he was stressed. He kept working as hard as he could to finish everything, and it paid off. His boss noticed how hard he was working and offered him a raise, as well as his choice in future projects. & \textbf{She had been devastated when the relationship ended and spent many empty days lying in bed, crying her eyes out and feeling that there was no point in going on. She hadn't even wanted to go to the party a couple of weeks later but a friend persuaded her. Whilst there, she locked eyes with a great looking guy and in no time they were chatting like old friends and exchanging numbers. Tomorrow is their second anniversary. }& The man had been in the job for ten years and had never had a payrise, despite his diligent hard work. One day he plucked up the courage to ask his boss for a rage and was curtly refused. That was it! He decided. No more working his hardest. He was going to slack as much as he could and not take any further pride in his work. \\ \hline
far, far & Bob designed a well-organized project management spreadsheet and loaded it to SharePoint for the other project managers to take a look at. Jack changed the formulas in the costing section. Rachel redesigned the charts. Tom hid several columns. Charlie reformatted cells. When Bob looked at the spreadsheet again, it was unusable. & She had been diagnosed with breast cancer but recovered. She felt surprised why she got the disease because she always leads a healthy life. She thinks that there is no point in worrying about the disease coming back. What she wants to do now is to improve her health condition and monitor her diet.  So she goes to gym every day and changes to a vegan diet. She feels more optimistic and does not worry about the disease anymore. & \textbf{Marsha was cooking dinner for the night.  She seasoned the food and left the room.  Her husband walked into the kitchen and saw the food on the stove.  He tasted it and added salt.  When Marsha walked back into the room, she remembered that she forgot to season the food, so she added salt.  The food was too salty and they had to order pizza that night.} \\ \hline
near, near & My daughter is away at college for the first time.  I am worried about her being so far from home and not near any family.  I am sure that she is having a great time and is too busy to check in with me.  I trust that she would contact me if something was wrong so she must be doing fine. & She worried every day about her son after he left for college. They had always been so close before. A month had gone by and she had not even had so much as a single phone call from him and they used to talk every day. It seemed he was so involved in his new life his mother just never occurred to him. & \textbf{As a concerned parent, my ultimate goal was to ensure my daughter's safety and well-being at college. With each passing day, I anxiously awaited news from her about her experiences and endeavors at the university. However, as the days turned into weeks and the weeks into months, the absence of troubling news brought me solace, reassuring me that she was thriving and happy in her new environment.} \\ \hline
near, far & Maria went abroad and she brings something to her friend Jucy a perfume as gift when she's back to her hometown and Jucy still not happy of the gift that Maria gives to her, she wants more than a perfume. Jucy should be content of what Maria gives to her at least Maria is thinking of her.  & \textbf{Once upon a time, two lifelong buddies stumbled upon a mysterious box in the middle of the forest. Curiosity piqued, they cautiously opened it and to their astonishment, found a rare and valuable treasure inside. Despite their better judgment, one of the buddies couldn't resist inspecting the gift closely, only to discover a small flaw that cast a shadow over their newfound joy. In the end, they both realized that their unexpected fortune should be cherished without questioning its imperfections.} & Although the friend really was innocent, he got arrested alongside his friends for being at the wrong place at the wrong time, but mostly because of his friends being the usual suspects. \\
\bottomrule
\end{tabular}}
\caption{
Examples of $ARN$ dataset covering four categories of far/near analogies facing far/near disanalogies. Correct answers are \textbf{boldfaced.}}
\label{tab:arn-examples}
\end{table*}

\section{Experimental Details}
\label{appendix:experimental-details}
The zero-shot and few-shot experiments on GPT3.5 and GPT4.0 were done using the OpenAI API, and each experiment took between 60 to 90 minutes, depending on the number of demonstrations we included in the prompt given to the model. The rest of the zero-shot experiments on FlanT5, Llama 2, Macaw, and UnifiedQA models were all executed on two NVIDIA Quadro RTX 8000 GPUs, each with 48 GB GPU memory. Zero-shot experiments on LLama 2 took around 60 minutes, and zero-shot experiments on FlanT5 and UnifiedQA took around 30 minutes each. Few-shot experiments done on Llama-2 were done using a batch size of 8 and took between 60 to 120 minutes to finish, depending on the number of demonstrations. Also, all the semantic similarities that were computed based on Sentence BERT \citep{reimers-2019-sentence-bert} were utilizing the all-mpnet-base-v2 pre-trained model.

\section{Prompts}
\label{appendix:prompts}

\subsection{Extracting Narrative Elements}
\label{appendix:extracting-narratives-elemenets}
We utilized GPT3.5 to extract all the elements mentioned in \autoref{subsec:elements-extraction} from narratives except for their corresponding proverbs that was already provided in the ePiC dataset. We tested both zero-shot extraction and few-shot extraction using solved demonstrations. Since using few-shot demonstrations yielded more consistent outputs, we extracted all the elements in the few-shot setting. The prompt we used is as follows: \textit{Return a list of \{element\} in the given narrative}. All the prompts were accompanied by three randomly solved demonstrations with narratives and their extracted elements that were extracted manually by the authors.

\subsection{Llama 2, GPT3.5, and GPT4.0 Evaluation}
The prompts given to Llama 2, GPT3.5, and GPT4.0 models for the main zero-shot and few-shot experiments consisted of multiple components:
\begin{enumerate*}
    \item the component in which the task is defined and some details are provided about the template that the model should use to return the answers;
    \item the component in which the demonstrations are included in the prompt (shown in a lighter color), which is not present in the zero-shot setting and is repeated multiple times in the few-shot setting depending on the specified number of demonstrations. 
    \item the component that shows the question that the model must solve:
\end{enumerate*}

\noindent\textit{narratives can be mapped to each other in terms of the high-level message they strive to convey. These high-level messages can be related to traditions, common knowledge, or moral principles. We call this mapping analogical mapping. Which one of the two narratives (1, 2) can create a better analogical mapping with the query narrative? Answer in the template: \{\{narrative\_x, because narrative\_x and query\_narrative are ...\}\}\\
\textcolor{gray}{---------\\
query\_narrative: \{query narrative demonstration\}\\
---------\\
narrative\_1: \{first candidate demonstration\}\\
---------\\
narrative\_2: \{second candidate demonstration\}\\
\#\#\#\#\#\#\#\#\#\#\\
\{Answer provided with its corresponding explanation\}}\\
---------\\
query\_narrative: \{query\_narrative\}\\
---------\\
narrative\_1: \{first candidate\}\\
---------\\
narrative\_2: \{second candidate\}\\
\#\#\#\#\#\#\#\#\#\#\\
narrative\\}

\subsection{Macaw, UnifiedQA, and FlanT5 Evaluation}
The prompt given to Macaw, UnifiedQA, and FlanT5 models that were utilized in zero-shot experiments was different from the prompt used for GPT3.5, GPT4.0, and Llama 2 models, because of the specific instruction template that was used in their fine-tuning. The content of the prompt is the same as the other prompt; however, to adopt a question-answering setting discussed in Macaw \citep{tafjord2021general}, UnifiedQA \citep{khashabi2020unifiedqa}, and FlanT5 \citep{chung2022scaling}, we slightly modified the prompt as follows:

\noindent\textit{narratives can be mapped to each other in terms of the high-level message they strive to convey. This high-level message can be related to traditions, common knowledge, or moral principles. We call this mapping analogical mapping. Which one of the two narratives (a, b) can create a better analogical mapping with the query narrative? query\_narrative: \{query narrative\}. \textbackslash \textbackslash n (a) \{first candidate\} (b) \{second candidate\}}  

\subsection{Generating Near Analogies}
\label{appendix:generating-near-analogies}
As it was mentioned in \autoref{subsec:selecting-analogies}, near analogies were created using the strengths of both LLMs and human supervision, ensuring that near analogies and their corresponding distractors form the same lower-order mappings with the query narratives while reflecting a given proverbial message. 

Since the lower-order mappings between distractors and query narratives are not based on exact lexical overlap and are based on semantic similarity, we tried to simulate the same environment for the lower-order mappings formed between near analogies and query narratives. Hence, when selecting the narrative elements that would form lower-order mappings in near analogies, synonyms of those particular elements were used to prompt LLM to generate narratives. We used a simple prompt to retrieve the synonyms as presented below:

\noindent\textit{shortly paraphrase or give a short synonym for \{element content\} without any explanations.}

The complete prompt that we used to generate the near analogies given the proverb and the elements that must be contained in the narrative is presented below:

\noindent\textit{Generate a random narrative with 4-8 sentences that uses \{elements type and their content\}. THE NARRATIVE MUST HAVE THE SAME CONCLUSION AS THE PROVERB: \{given proverb\}. Only return the narrative without any explanation, and also DO NOT mention the proverb. }

\section{Additional Discussion}
\label{appendix:additional-experiments}

\begin{figure}[h]
    \centering
    \includegraphics[width=0.7\linewidth]{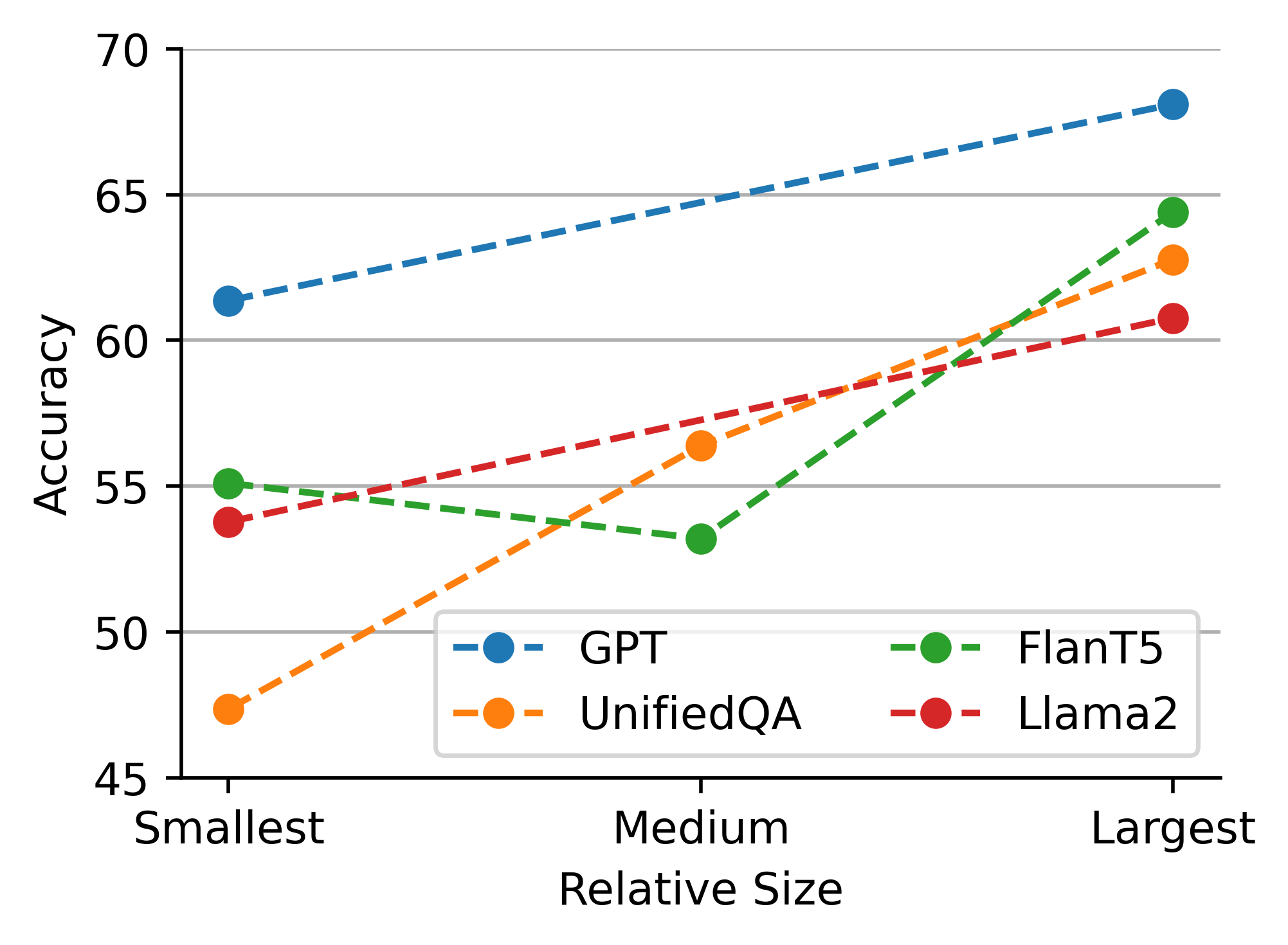}
    \caption{Effect of scaling each family of models up on the analogical reasoning performance in a zero-shot setting.}
    \label{fig:effect-of-size}
\end{figure}

\subsection{Effect of Scaling Up the Models' Size on Analogical Reasoning}
\autoref{fig:effect-of-size} demonstrates the zero-shot performance of all models on $ARN$ with different numbers of parameters. In line with the prior finding \citep[e.g.,][]{wei2022emergent}, we found that scaling up the models enhances the analogical reasoning capabilities of models. The increase in the performance of models was 10 absolute points, on average, with UnifiedQA gaining the most from scaling up the number of parameters, as high as almost 20 absolute points.

\begin{figure}[h]
    \centering
    \includegraphics[width=1.0\linewidth]{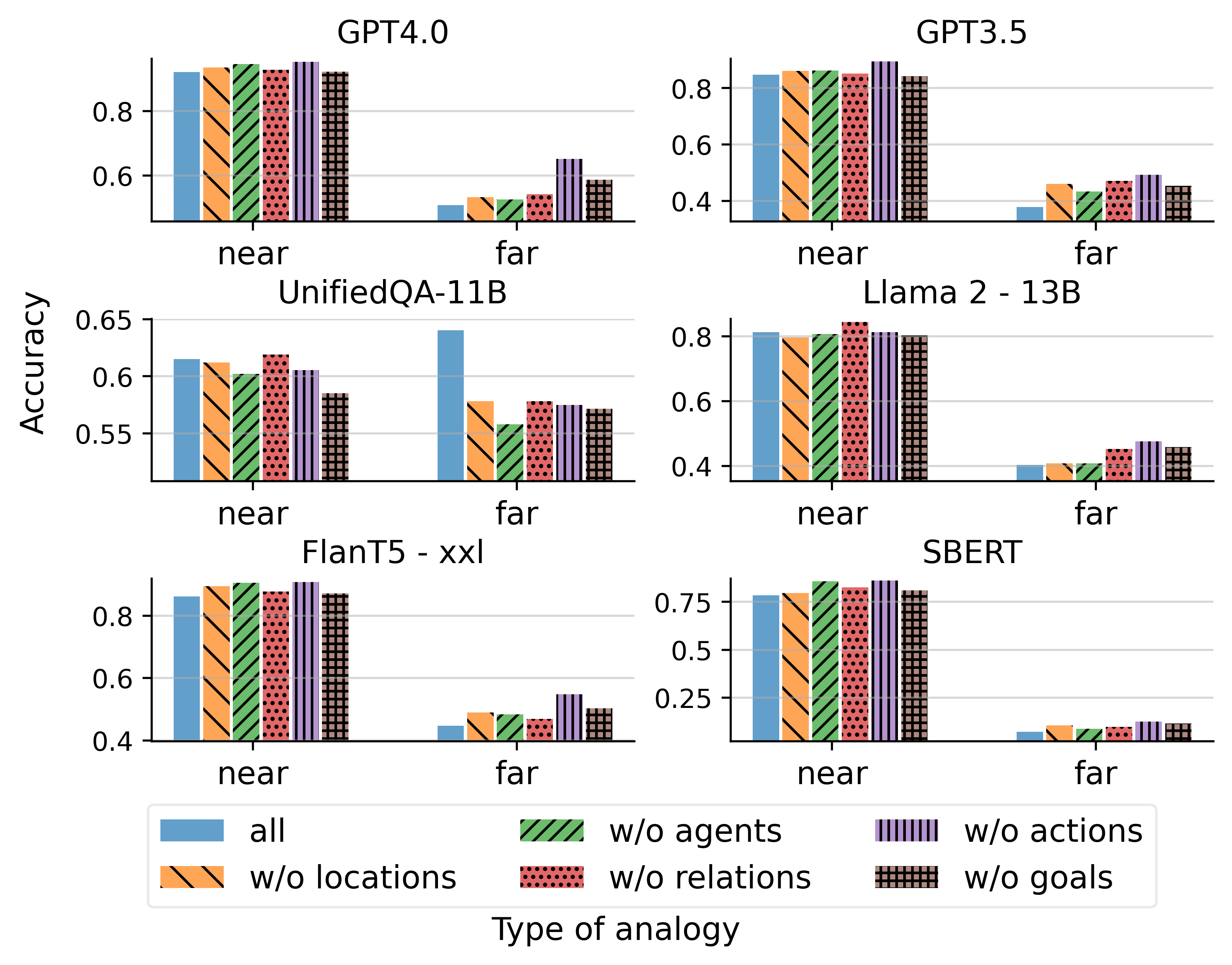}
    \caption{Performance of all models on $ARN$, in a zero-shot setting, categorized jointly by the type of analogies and distractors.}
    \label{fig:effect-of-contextual-dimensions-distractors-all}
\end{figure}

\subsection{Distracting Effect of Surface Mappings}
\label{appendix:Distracting-Effect-of-Surface-Mappings}
Performance of all models on $ARN$ categorized by the type of analogies, disanalogies, and specific surface mappings included in distractors, is demonstrated in \autoref{fig:effect-of-contextual-dimensions-distractors-all}. We observed that for all the models except for UnifiedQA-11B, including all types of distractors is more distracting than scenarios where some are excluded, especially in far analogies. In other words, each distractor has its own individual effect on the performance and their aggregated effect is stronger than individuals. Also, among different elements, the lower-order mappings that were formed based on the actions of narratives had the most distracting effect on the analogical reasoning abilities of models. However, we saw a different trend for UnifiedQA-11B, where including all the distractors did not yield the most distracting effect, and indeed the model had its highest performance when all types of distractors were included. 

\begin{table*}[!ht]
\centering
\small
\begin{tabular}{p{2.5cm}|c|c|c|c|c}
\toprule
\thead{Task Partition} & \thead{(near, far)} & \thead{(near, near)} & \thead{(far, far)} & \thead{(far, near)} & \thead{Avg.} \\
\midrule
GPT3.5 & 88.1 (81.9) & 81.3 (74.3) & 50.4 (50.2) & 21.7 (25.3) & 60.3 (57.9) \\
GPT4.0 & 94.0 (92.7) & 92.5 (89.0) & 57.1 (54.1) & 29.1 (17.7) & 68.1 (63.4) \\
UnifiedQA-large & 43.2 (46.9) & 49.1 (52.2) & 47.2 (46.2) & 50.5 (47.1) & 47.5 (48.1) \\
UnifiedQA-3B & 66.4 (61.0) & 68.4 (62.7) & 47.3 (47.3) & 44.4 (38.9) & 56.6 (52.5) \\
UnifiedQA-11B & 60.7 (63.4) & 61.2 (65.3) & 54.8 (53.7) & 74.6 (52.9) & 62.8 (58.8) \\
Llama2-7B & 63.4 (74.4) & 58.0 (66.0) & 50.1 (47.8) & 43.1 (36.8) & 53.7 (56.3) \\
Llama2-13B & 80.9 (82.3) & 81.4 (74.8) & 44.5 (45.2) & 35.4 (41.7) & 60.5 (61.0) \\
FlanT5-large & 84.5 (84.2) & 80.3 (79.6) & 41.4 (39.0) & 14.4 (11.9) & 55.1 (53.7) \\
FlanT5-xl & 78.9 (74.5) & 68.3 (67.0) & 44.7 (42.0) & 21.1 (20.4) & 53.2 (51.0) \\
FlanT5-xxl & 89.9 (76.1) & 81.3 (72.0) & 51.1 (48.0) & 35.6 (34.3) & 64.4 (57.6) \\
Macaw-11B & 88.0 (86.4) & 84.6 (82.3) & 42.1 (42.0) & 35.8 (34.0) & 62.8 (61.2) \\
\bottomrule
\end{tabular}
\caption{
Accuracy (with random baseline of $50\%$) of tested LLMs in zero-shot setting, on $ARN$, with and without (in parentheses) explicit instructions to attend to system mappings, in four categories of far/near analogies facing far/near disanalogies, e.g., (near, far) indicates the setting where near analogies are alongside far disanalogies.
}
\label{tab:results-with-similarity-instructions}
\end{table*}

\subsection{Do Models Find Analogies and Similarities Different From Each Other?}
\label{appendix:do-models-find-analogical-relatedness-and-similarities-different}

While in our main experiments, we explicitly prompted the model to find system mappings and mappings based on proverbial high-level messages, in an additional experiment, we prompted models to look for mere "similarity," with the performance of models in both settings demonstrated in \autoref{tab:results-with-similarity-instructions}. We observed that the performance with the explicit instruction to look for analogical mappings is higher compared to the setting without explicit instructions, which is reasonable given the nature of the task and models following the instructions. The relatively small gap between the two settings might be due to the models seeing analogical mappings as more sound compared to surface mappings, which is similar to humans \citep{gentner1993roles}. These results suggest that the low performance of the models in \autoref{tab:main-results} is not due to the chosen prompt. However, future work can investigate more effective approaches to activate the analogical reasoning abilities of LLMs. 

\begin{table*}[ht]
\centering
\small
\scalebox{0.8}{
\begin{tabular}{ p{0.30\textwidth} | p{0.26\textwidth} | p{0.30\textwidth} | p{0.20\textwidth}}
\toprule
 \textbf{Query narrative} & \textbf{narrative 1} & \textbf{narrative 2} & \textbf{Model prediction} \\
\midrule
 The bees often pollinate the flowers in the early morning, when the dew rests upon the blades of grass and the birds are still finding their tune. The uniform bloom of the flowers allowed for a uniform death. One bee decided to go against the grain one day and pollinate when the birds were asleep in the dead of night. While at first glance the flower seemed unaffected, it began to wilt at an accelerated pace compared to those who had been pollinated the next morning. It was a fatal mistake. & Not everybody was paying attention to our pastor when we were at the Football tournament. Pastor took more than two hours trying to preach. but from the observation no one was interested in his sermon, his preaching is liked by many, but not at this particular moment that we came to play and have fun. & A man named Toribo carried a bag full of charms over the mountain to the fairest princess to convince her he was nice enough for her to love him. He thought that charms of flattery would make her victim to his control, but when he found the Princess she was uninterested by his fake, flowery words. Instead of committing herself to Toribo, the princess fell in love with herself and chose to wait for a person who loved her as she was, and not only someone who threw superficial flatteries her way. & narrative\_2, because narrative\_2 and query\_narrative both convey the message that going against the natural or expected order of things can lead to unintended negative consequences. \\
\midrule
 Robbie delighted in telling people stories about his wealthy background even though he actually came from a poor home.  In college, his new roommate confronted Robbie about his true background.  His roommate told Robbie he also told people false stories about his rich parents.
 &  When we began our charlity organization at Mathare slums, which was aiming at finding the young people who had gone into crimes and rehabilitate them we started by making friendship with those street children at the slums. Finally we were able to track and talk to the young gangsters for we were able to locate where they used to hide by the help of the one we friended.
 & Upon clearing college, I had big ambitions. I wanted to take over from my father as the CEO of our family's companies. My father admonished me saying that I needed to go get a few professional courses on business management before he could hand over the reins to me. At first I was hesitant to go back to school, but I saw the sense and reckoned that I needed to tool myself for the duties that awaited me. There was a process.
 & narrative\_2, because narrative\_2 and query\_narrative are both about individuals who have to confront the reality of their situations and make adjustments accordingly. \\
\midrule
Max was a ferocious looking and sounding canine.  People walking by him would cross to the other side of the road.  One day a deaf man was walking by the yard and Max was going off with his incessant yowling.  Max ran to the sidewalk, still yepping, but let the man walk by unscathed. & In response to being ignored by her so-called friend at the party, she decided to balance things out by not responding to her messages the next day.  She later found out that her friend's boyfriend had left her and that was why she was so quiet at the party. She felt terrible for ignoring her messages the next day and wished she had not responded in this negative way. & Billy was so excited to win the race for class president. He worked so hard to win, and he was so proud of himself. His opponent wasn't so happy. He kept telling Billy that he was going to make sure he failed as president, and that everyone would be sorry they voted for him. But considering that he did the same thing last year, Billy knew it was all talk. & narrative\_1, because narrative\_1 and the query\_narrative both involve a misunderstanding due to lack of information, leading to regrettable actions or reactions. \\
\bottomrule
\end{tabular}}
\caption{
Examples of GPT4.0 errors alongside the free-text rationales generated by the model in a zero-shot setting.}
\label{tab:errors}
\end{table*}

\end{document}